\documentclass[10pt,twocolumn,letterpaper]{article}

\usepackage{cvpr}              % To produce the CAMERA-READY version
\usepackage{graphicx}
\usepackage{amsmath}
\usepackage{amssymb}
\usepackage{booktabs}
\usepackage{algorithm}
\usepackage{algorithmic}
\usepackage{makecell}
\usepackage{multirow}
\usepackage[pagebackref,breaklinks,colorlinks]{hyperref}

\usepackage[accsupp]{axessibility}  % Improves PDF readability for those with disabilities.
\usepackage[capitalize]{cleveref}
\crefname{section}{Sec.}{Secs.}
\Crefname{section}{Section}{Sections}
\Crefname{table}{Table}{Tables}
\crefname{table}{Tab.}{Tabs.}

\def\cvprPaperID{7499} % *** Enter the CVPR Paper ID here
\def\confName{CVPR}
\def\confYear{2023}

\begin{document}

%%%%%%%%% TITLE - PLEASE UPDATE
\title{Model Barrier: A Compact Un-Transferable Isolation Domain\\
for Model Intellectual Property Protection} 

\author{Lianyu Wang$^1$\footnotemark[1], \; Meng Wang$^2$\footnotemark[1], \; Daoqiang Zhang$^1$\footnotemark[2], \; Huazhu Fu$^2$\footnotemark[2] \\
\small{$^1$College of Computer Science and Technology, Nanjing University of Aeronautics and Astronautics, China.}\\ 
\small{$^2$Institute of High Performance Computing (IHPC), Agency for Science, Technology and Research (A*STAR), Singapore 138632.}
}

\maketitle

%%%%%%%%% ABSTRACT
\begin{abstract}
As scientific and technological advancements result from human intellectual labor and computational costs, protecting model intellectual property (IP) has become increasingly important to encourage model creators and owners. Model IP protection involves preventing the use of well-trained models on unauthorized domains.
To address this issue, we propose a novel approach called Compact Un-Transferable Isolation Domain (CUTI-domain), which acts as a barrier to block illegal transfers from authorized to unauthorized domains.
Specifically, CUTI-domain blocks cross-domain transfers by highlighting the private style features of the authorized domain, leading to recognition failure on unauthorized domains with irrelevant private style features.
Moreover, we provide two solutions for using CUTI-domain depending on whether the unauthorized domain is known or not: target-specified CUTI-domain and target-free CUTI-domain. Our comprehensive experimental results on four digit datasets, CIFAR10 \& STL10, and VisDA-2017 dataset demonstrate that CUTI-domain can be easily implemented as a plug-and-play module with different backbones, providing an efficient solution for model IP protection.
\end{abstract}

\renewcommand{\thefootnote}{\fnsymbol{footnote}} %将脚注符号设置为fnsymbol类型，即特殊符号表示
\footnotetext[1]{L.~Wang and M.~Wang contributed equally to this work.} %对应脚注[1]
\footnotetext[2]{Corresponding author: D.~Zhang (dqzhang@nuaa.edu.cn) and H.~Fu (hzfu@ieee.org).} %对应脚注[2]

%%%%%%%%% BODY TEXT
\section{Introduction}
\label{sec:intro}

\begin{figure}[!t]
  \centering
   \includegraphics[width=0.9\linewidth,trim=190 70 190 70,clip]{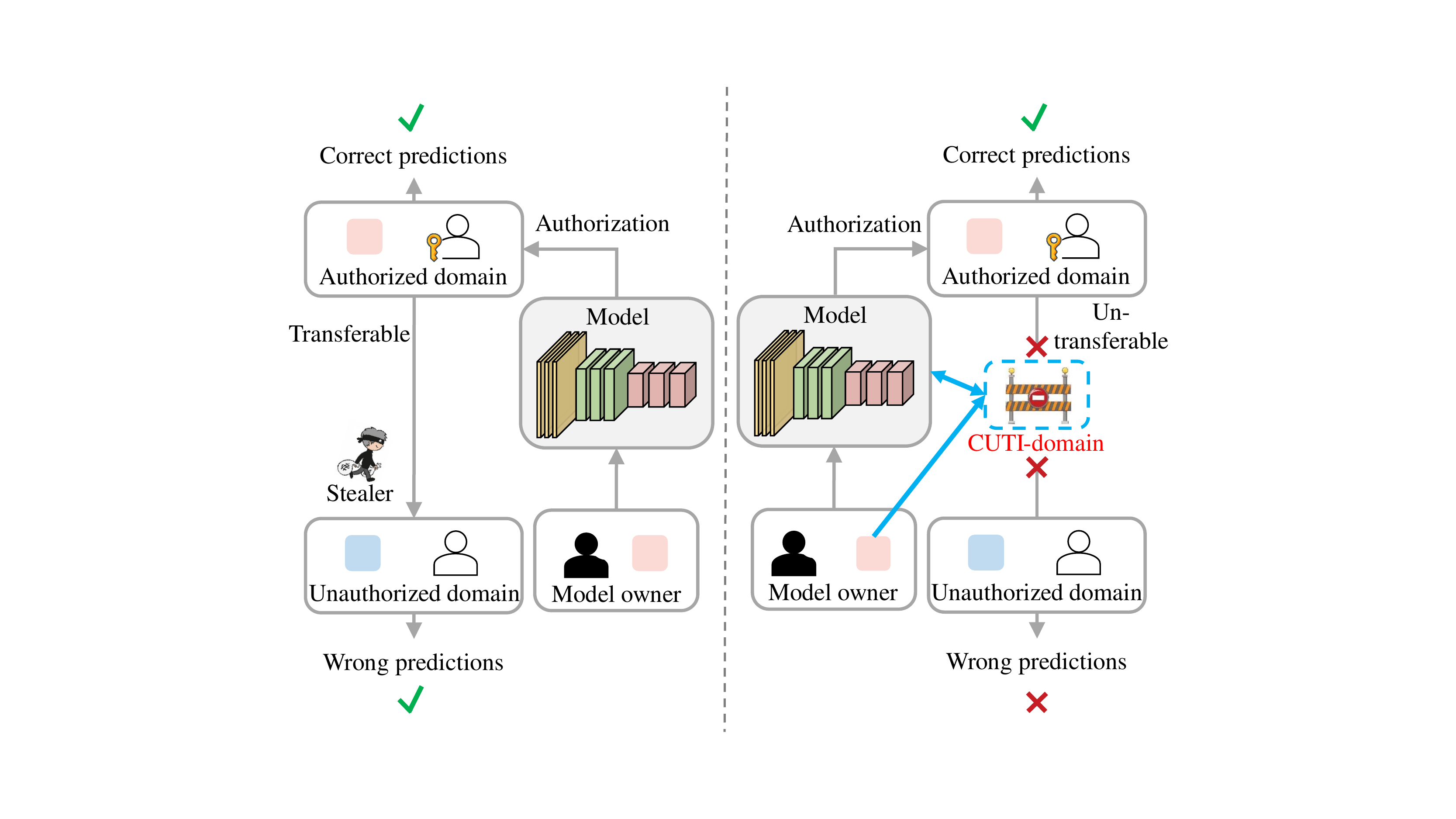}
   \caption{Model IP protection with our proposed CUTI-domain. \textbf{Left}: In standard supervised learning (SL), the model owner trains a high-performance model on the authorized domain (pink square) and then authorizes a specific user. Authorized users have the right to use the model on the authorized domain to get the correct prediction. However, a stealer can easily access the model on the unauthorized domain (blue square), which violates the legitimate rights and interests of the model owner. 
   \textbf{Right}: Our method constructs a CUTI-domain between the authorized and unauthorized domains, which could block the illegal transferring and lead to a wrong prediction for unauthorized domains.}
   \label{figure1}
   \vskip -10pt
\end{figure}

% Deep learning has achieved remarkable performance on various vision tasks, including image recognition~\cite{ref1, ref2}, semantic segmentation~\cite{ref3}, and object localization~\cite{ref4}, etc. 
The recent success of deep learning models heavily relies on massive amounts of high-quality data, specialized training resources, and elaborate manual fine-tuning~\cite{LeCun2015,ref5, Bommasani2021,Shamshad2022}. 
Obtaining a well-trained deep model is both time-consuming and labor-intensive~\cite{ref8}. 
Therefore, it should be protected as a kind of scientific and technological achievement intellectual property (IP)~\cite{ref29,ref42}, thereby stimulating innovation enthusiasm in the community and further promoting the development of deep learning. 
As shown in Fig.~\ref{figure1}, in supervised learning (SL), the model owner uses the overall features of the authorized domain (pink square) for training, obtains a high-performance model, and grants the right to use it to a specific user. Authorized users can use the model on the authorized domain to obtain correct predictions. However, since the model is trained with overall features, its potential feature generalization region is large and may cover some unauthorized domains. Therefore, there is a natural pathway between the authorized domain and the unauthorized domain, and the released high-performance model obtained by SL can be illegally transferred to the unauthorized domain (blue squares) through methods such as domain adaptation~\cite{ref15,ref16}, and domain generalization~\cite{ref17,ref18}, to obtain correct prediction results. This presents a challenge in protecting well-trained models. One of the most concerning threats raised is “Will releasing the model make it easy for the main competitor to copy this new feature and hurt owner differentiation in the market?” Thus, the model IP protection has been proposed to defend against model stealing or unauthorized use.

% This presents a challenge in protecting well-trained models.
% That said, it's not difficult to \textit{\textbf{steal}} a well-trained model.

A comprehensive intellectual property (IP) protection strategy for deep learning models should consider both ownership verification and applicability authorization~\cite{ref9,Wang2022}.
Ownership verification involves verifying who has permission to use the deep model by embedding watermarks~\cite{ref10, ref11}, model fingerprint~\cite{ref12}, and predefined triggers~\cite{ref13}. The model owner can grant usage permission to a specific user, and any other users will be infringing on the owner's IP rights. However, an authorized user can easily transfer the model to an unauthorized user, so the model owner must add special marks during training to identify and verify ownership.
Moreover, these methods are vulnerable to fine-tuning, classifier retraining, elastic weight consolidation algorithms, and watermark overwriting, which can weaken the model's protection.
On the other hand, applicability authorization involves verifying the model's usage scenarios. Users with permission can apply the deep model for the tasks specified by the model owner, and it is an infringement to use it for unauthorized tasks~\cite{ref9}. However, users can easily transfer high-performance models to other similar tasks to save costs, which is a common and hidden infringement. Therefore, if the performance of the model can be limited to the tasks specified by the owner and reduced on other similar tasks, unauthorized users will lose confidence in stealing and re-authoring the model.
To achieve this, a non-transferable learning (NTL) method is proposed~\cite{ref9}, which uses an estimator with a characteristic kernel from Reproducing Kernel Hilbert Spaces to approximate and increase the maximum mean difference between two distributions on finite samples. However, the authors only considered using limited samples to increase the mean distribution difference of features between domains and ignored outliers. The convergence region of NTL is not tight enough. Moreover, the calculation of the maximum mean difference is class-independent, which reduces the model's feature recognition ability in the authorized domain to a certain extent.

To address the challenges outlined above, we \textbf{first} propose a novel approach called the Compact Un-Transferable Isolation (CUTI) domain to prevent illegal transferring of deep models from authorized to unauthorized domains.  
Our approach considers the overall feature of each domain, consisting of two components: shared features and private features. Shared features refer to semantic features, while private features include stylistic cues such as perspective, texture, saturation, brightness, background environment, and so on. We emphasize the private features of the authorized domain and construct a CUTI-domain as a model barrier with similar private style features. This approach prevents illegal transfers to unauthorized domains with new private style features, thereby leading to wrong predictions. 
\textbf{Furthermore}, we also provide two CUTI-domain solutions for different scenarios. When the unauthorized domain is known, we propose the target-specified CUTI-domain, where the model is trained with a combination of authorized, CUTI, and unauthorized domains. When the unauthorized domain is unknown, we use the target-free CUTI-domain, which employs a generator to synthesize unauthorized samples that replace the unauthorized domain in model training.
\textbf{At last}, our comprehensive experimental results on four digit datasets, CIFAR10 \& STL10, and VisDA-2017 demonstrate that our proposed CUTI-domain effectively reduces the recognition ability on unauthorized domains while maintaining strong recognition on authorized domains. Moreover, as a plug-and-play module, our CUTI-domain can be easily implemented within different backbones and provide efficient solutions.\footnote[1]{\textcolor{red}{https://github.com/LyWang12/CUTI-Domain.}}

\section{Related Work}
\subsection{Model IP Protection}
There are currently two main categories of methods for IP protection, including ownership verification and applicability authorization. 
For ownership verification, the most classic method is watermarking embedding~\cite{ref23}. Kuribayashi \etal~\cite{ref24} proposed a quantifiable watermark embedding method, which reduces the variation caused by embedding watermarks. Adi~\etal~\cite{ref25} proposed a tracking mechanism in a black-box way. However, such watermark embedding approaches have been proved to be susceptible to some watermark removal and watermark overwriting methods. In our experiments, a simple watermark is embedded into the model for ownership verification by triggering mis-classification. Comprehensive experimental results demonstrate that the proposed CUTI-domain is resistant to the common watermark removal methods.

Applicability authorization is derived from usage authorization. Usage authorization usually uses a preset private key to encrypt the whole/part of the network. Only authorized users can obtain the private key and then use the model. There are many advanced methods for usage authorization. For example, Alam~\etal~\cite{ref28}. proposed an explicit locking mechanism for a lightweight deep neural network, utilizing S-Boxes with good cryptographic properties to lock each training parameter of a DNN model. Without knowledge of the legitimate private key, unauthorized access can severely degrade the accuracy of the model. Song~\etal~\cite{ref30} analyzed and calculated the critical weight of the deep neural network model, and significantly reduced the time cost by encrypting the critical weight to lock the deep neural network model against unauthorized use. 
Wang~\etal~\cite{ref9} proposed data-based applicability authorization named NTL, which
preserves model performance on authorized data while degrading model performance in other data domains.  
Compared to the above, we construct a new class-dependent CUTI-domain with infinite samples whose features are more similar with the source domain. By decreasing the performance of the model on the CUTI-domain and the target domain, the generalization bound of the model can be compacter, thereby constraining the model performance within the authorized source domain.

\subsection{Domain Transferring}
In practice, domain gaps may arise due to different data collection scenarios in different domains. Domain adaptation (DA) and domain generalization (DG) are common solutions used to alleviate domain gaps~\cite{WANG2018135,9847099}.
DA refers to transferring a model from a labeled source domain to an unlabeled but relevant target domain where the target domain's data is accessible during the training process~\cite{ref14}. DG differs from DA in that the target domain is inaccessible during model training~\cite{ref44,ref45}. 
DANN~\cite{ref116} s a classic DA method that introduced a gradient inversion layer and a domain discriminator to confuse the feature distributions of the two domains. Subsequently, CDAN~\cite{ref117} further introduced categorical information entropy into the domain discriminator to alleviate the class mismatch problem. 
For DG, Tobin~\etal~\cite{ref132} used domain randomization to generate more training data from simulated environments for generalization in real environments. Prakash~\etal~\cite{ref130} further considered the structure of the scene when randomly placing objects for data generation, enabling the neural network to learn how to utilize context when detecting objects.
Recently, some methods have been proposed and successfully applied to cross-domain applications by seeking an intermediate state between the source domain and the target domain, emphasizing the similarity between domains to improve the model's transferability~\cite{ref46,ref47,ref48,ref49,ref50}. 
In contrast, this paper aims to seek an intermediate state to highlight the difference between the authorized source domain and the unauthorized target domain, constraining the feature transferability of the model and protecting the scientific and technological achievements IP of the model owner.

\section{Methodology}
In this section, we first present our proposed CUTI-domain with the aim of developing a solution that limits the performance of the model to authorized source domains and reduces feature recognition capabilities on the unauthorized target domain. Then, depending on whether the unauthorized target domain is known or not, we provide two solutions, target-specified CUTI-domain and target-free CUTI-domain, for protecting the model IP. 

\subsection{Compact Un-Transferable Isolation Domain}
In the deep neural network model, the overall features extracted by the feature extractor include two abstract components, \textit{i.e.}, semantic features, and style features. Semantic features reflect the structural information of samples and play a leading role in sample recognition; while style feature refers to a series of weakly related clues, such as perspective, texture, saturation, brightness, and background environment. For different domains with the same task, semantic features are shared, while style features are private. Most of the previous DA and DG works have been devoted to improving feature transferability between domains, \textit{i.e.}, strengthening the focus of the model on shared features while suppressing seemingly disturbing private style features. However, to protect the intellectual property of the model, this paper aims to limit the feature recognition ability of the model by highlighting private style features of the source domain through style transfer, thus leading to the failure of recognition on target domains that contain irrelative private style.

Style transfer techniques suggest that styles are homogeneous and composed of repeated structural motifs. Two images can be considered similar in style if the features extracted by a trained classifier share the same statistics~\cite{ref33,ref31}. First- and second-order statistics are often used as style features due to their computational efficiency. These statistics refer to the mean and variance of the extracted features. On the other hand, semantic features only contain pure semantic information and exclude any style information. Similar to~\cite{ref32}, the semantic feature \(f_s\) of the extracted feature \(f\) can be obtained by removing the style features, as follows: 
\begin{equation} 
f_s = \frac{{f - {\mu(f)}}}{{{\sigma(f)}}},
\end{equation} 
where \(\mu(f)\) and \(\sigma(f)\) denote the mean and variance of \(f\). Furthermore, style can be re-assign by \(f_s \cdot \gamma  + \beta\), where \(\gamma\) and \(\beta\) are learned parameters. Afterward, Huang~\etal~\cite{ref33} further explored adapting \(f\) to arbitrarily given style by using style features of another extracted feature instead of learned parameters. 

\begin{figure}[!t]
  \centering
   \includegraphics[width=0.8\linewidth,trim=220 120 220 120,clip]{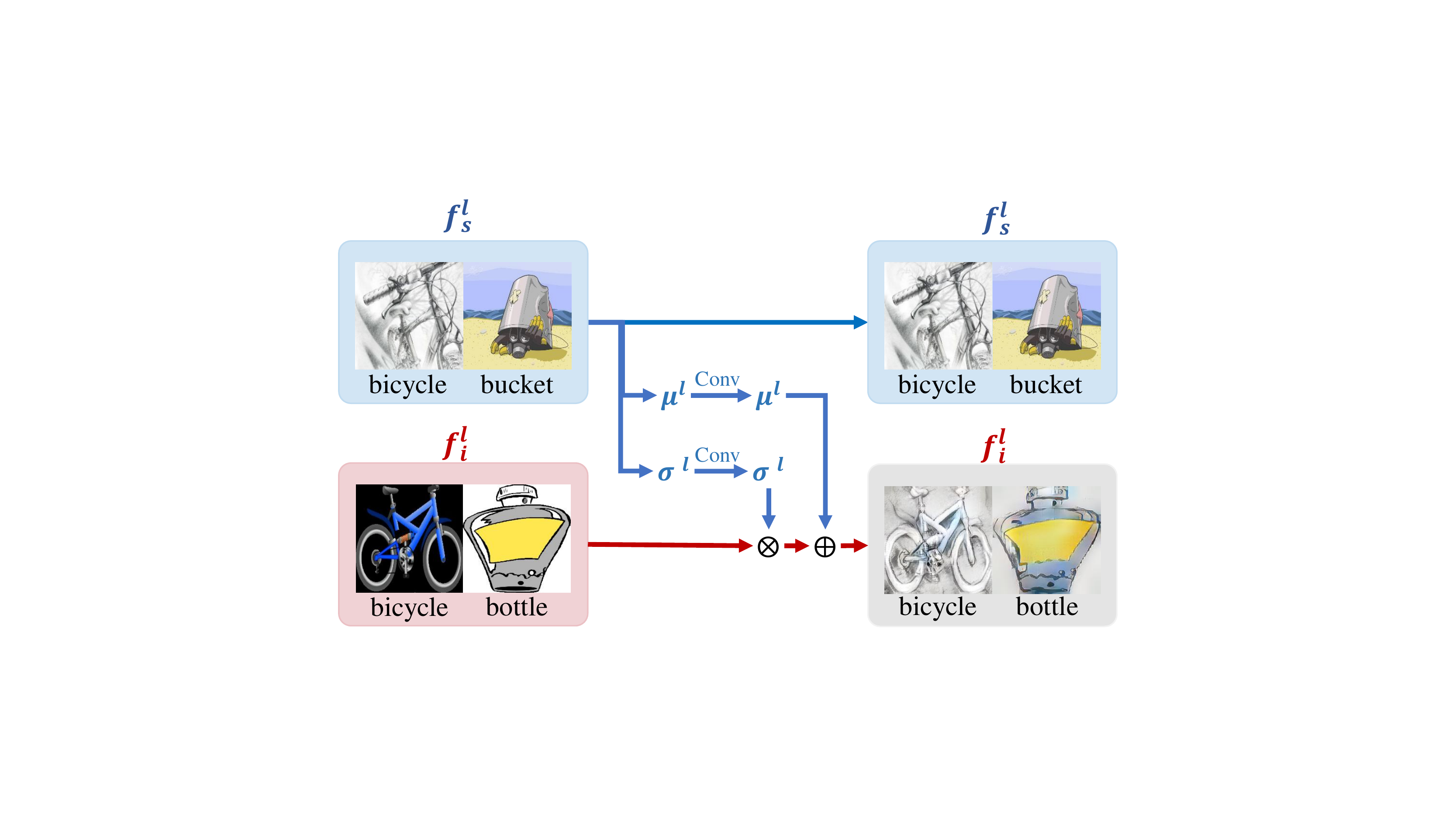}
   \caption{Illustration of our proposed CUTI-domain generator. The output feature of the source domain \(f_s^l\) and the CUTI-domain \(f_i^l\) in the \(i\)-th feature extractor block are sent to the CUTI-domain generator, and then the mean \(\mu^l\) and variance \(\sigma^l\) of \(f_s^l\) is fused with \(f_i^l\), so that the private style features of the updated \(f_i^l\) are closer to the feature of source domain \(f_s^l\) while the original semantics features are maintained.}
   \label{figure2}
   \vskip -10pt
\end{figure}

% Inspired by them, a novel CUTI-domain is designed. The style features of the source domain are randomly extracted and fused with the overall features of the CUTI-domain to make the private style features of the CUTI-domain closer to the source domain, and then reduce the feature recognition ability of the model on the CUTI-domain and the target domain to implicitly block the pathway between the source and the target domain, thus compact the performance of the model to the source domains.

Drawing inspiration from these ideas, we propose a novel CUTI-domain design that incorporates style features from the source domain. Specifically, we randomly extract style features from the source domain and fuse them with the overall features of the CUTI-domain, making the private style features of the CUTI-domain more similar to those of the source domain. By doing so, we reduce the model's ability to recognize features on both the CUTI-domain and the target domain, thereby implicitly blocking the pathway between the source and target domains and limiting the model's performance to the source domains.

Our CUTI-domain is generated by the CUTI-domain generator, which is a lightweight, and plug-and-play module, as shown in Fig.~\ref{figure2}. \(f_s^l\) and \(f_i^l\) represent the deep features of the \(i\)-th feature extractor block in the source domain and the CUTI-domain, respectively. First, \(f_s^l\) and \(f_i^l\) are sent into CUTI-domain generator in parallel, and then the mean \(\mu^l\) and variance \(\sigma^l\) of \(f_s^l\) are calculated according to the channel as private style features, followed by a \(1 \times 1\) convolution layer \(Conv\). Next, the \(\mu^l\) and \(\sigma^l\) are multiplied and added channel-wisely by \(f_i^l\) as: 
\begin{equation}
\label{eq2}
f_i^l \leftarrow (f_i^l \bigotimes Conv(\sigma ^l)) \bigoplus Conv(\mu ^l).
\end{equation} 
As can be seen in Fig.~\ref{figure2}, the private style features of updated \(f_i^l\) are closer to those of \(f_s^l\), while retaining its original semantic features. Through continuous computation, CUTI-domain generators can construct a labeled CUTI-domain containing a similar private style to the source domain.

\subsection{Model IP Protection with CUTI-domain}

\subsubsection{Target-Specified CUTI-Domain}

\begin{figure}[!t]
  \centering
   \includegraphics[width=0.9\linewidth,trim=235 160 235 140,clip]{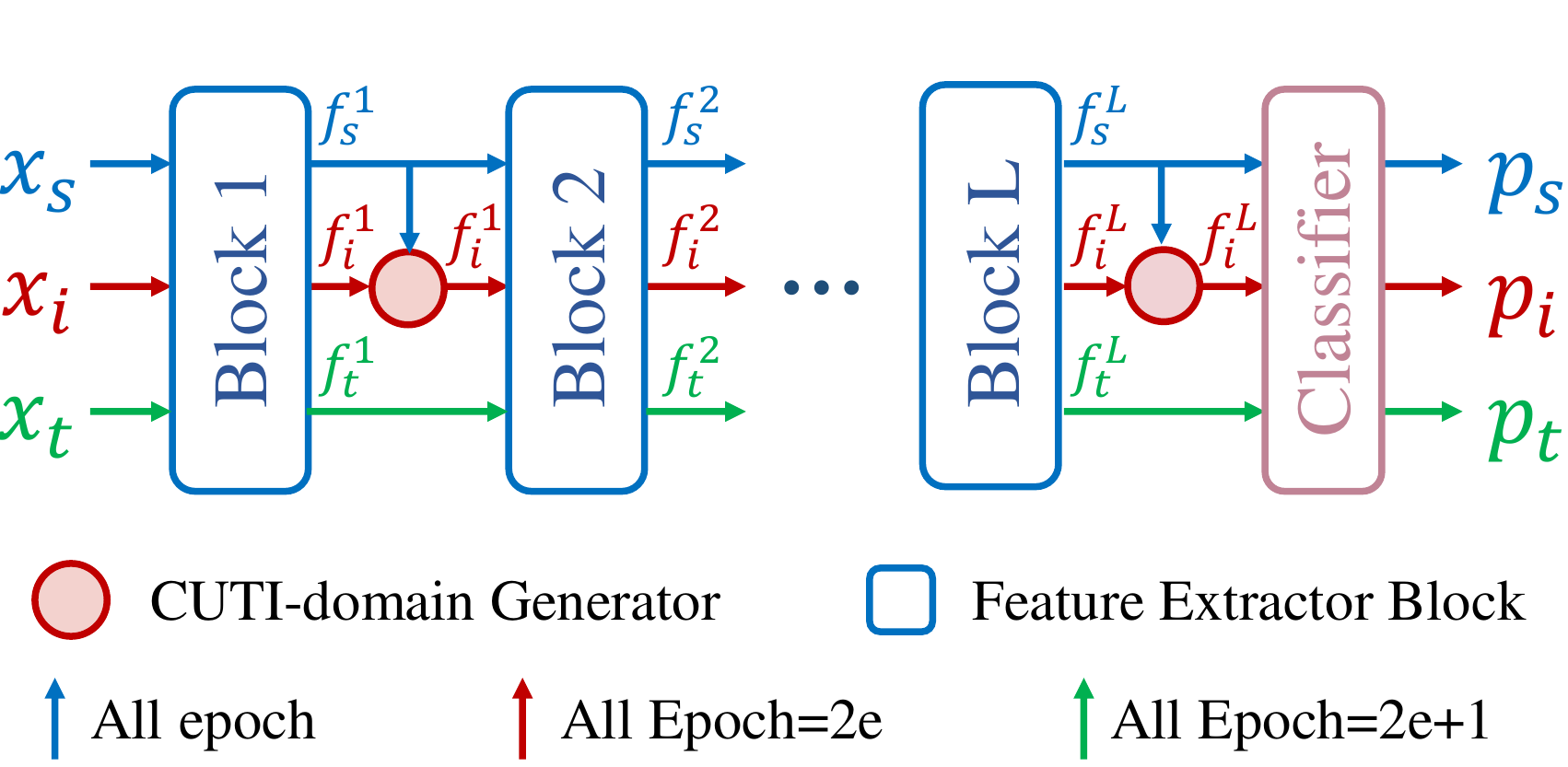}
   \caption{Illustration of the framework trained with our proposed CUTI-domain. The whole framework consists of \(L\) feature extractor blocks, \(L\) CUTI-domain generators, and a classifier. When the epoch is equal to an even/odd number, samples of the source domain \(x_s\) and CUTI-domain \(x_i\) / target domain \(x_t\) are fed into the feature extractor block in parallel. \(p_s\), \(p_i\) and \(p_t\) denote their prediction results.}
   \label{figure3}
   \vskip -10pt
\end{figure}

We first introduce how to utilize our CUTI-Domain with a given unauthorized target domain.
Fig.~\ref{figure3} illustrates the whole framework trained with our proposed CUTI-domain, which consists of \(L\) feature extractor blocks, \(L\) CUTI-domain generators and a classifier. \(x_s\), \(x_i\) and \(x_t\) denote the data of the source domian, CUTI-domain and target domain respectively.
During training, when \(epoch=2e\), \textit{i.e.}, epoch is equal to an even number, \(x_s\) and \(x_i\) are fed into the feature extractor blocks in parallel, followed by a CUTI-domain generator. The classifier at the end of the network is used to predict the category of the sample, and the prediction results are denoted by \(p_s\) and \(p_i\), respectively. When \(epoch=2e+1\), \textit{i.e.}, epoch is equal to an odd number, \(x_s\) and \(x_t\) are input into the network parallelly without CUTI-domain generators, \(p_s\) and \(p_t\) denote their predicted results.

Based on the designed framework, an alternative loss function \(\cal L\) according to the epoch is utilized:
\begin{equation} 
\small
  {\cal L} = \begin{cases}
    KL(p_s ||y_s ) - KL(p_{\textcolor{red}{i}} ||y_{\textcolor{red}{i}} ), & \text{if} \; epoch = 2e, \\
    KL(p_s ||y_s ) - KL(p_{\textcolor{red}{t}} ||y_{\textcolor{red}{t}}), & \text{if} \; epoch = 2e+1,
  \end{cases} 
  \label{eq1}
\end{equation}
where \(KL(\cdot)\) stands for Kullback–Leibler divergence. By reducing the \(KL(\cdot)\) between prediction results and the labels on the source domain, and expanding the \(KL(\cdot)\) prediction results and the labels on the CUTI-domain/target domain, the feature recognition ability of the model in the source domain can be gradually improved, and the ability in the CUTI-domain/target domain can be gradually reduced, thus effectively constrain the performance of the model within authorized source domains for IP protection.
Finally, we summarize the strategy of our proposed target-specified CUTI-domain in Algorithm~\ref{alg:alg1}.

During training, we initialize the CUTI-domain with the target domain training set, and feed data of source domain training set \(x_s\), CUTI-domain \(x_i\), and target domain training set \(x_t\) into the network in parallel to train the model as in Algorithm~\ref{alg1}. At test time, model performance is evaluated on the source domain test set and the target domain test set. In an ideal situation, the model can maintain high sample recognition ability on the authorized source domain, but achieve poor performance on the unauthorized target domain. 

\begin{algorithm}[!t]
\caption{Target-Specified CUTI-Domain.}\label{alg:alg1}
\begin{algorithmic}[1]
\REQUIRE{The source domain \(x_s\), CUTI-Domain \(x_i\), the target domain \(x_t\), number of feature blocks \(L\), the model parameters $\theta$.}
\STATE $ \text{Initialize CUTI-domain with the target domain.} $
\STATE $ \textbf{For \(epoch=1\) to \(Max_{epochs}\) do} $
\STATE $ \textbf{\qquad If \textcolor{red}{\(epoch=2i+1\)} do}$
\STATE $ \text{\qquad \qquad Calculate the output of \(x_s\), \(x_i\) in the \(l\)-th} $
$ \text{\qquad \qquad feature block: \(f_s^l\), \(f_i^l\).} $
\STATE $ \textbf{\qquad \qquad For \(l=1\) to \(L\) do} $
\STATE $ \text{\qquad \qquad \qquad Update \(f_i^l\) according to Eq.~(\ref{eq2}).} $
\STATE $ \textbf{\qquad \qquad End For} $
\STATE $ \textbf{\qquad If \textcolor{red}{\(epoch=2i\)} do} $
\STATE $ \text{\qquad \qquad Calculate the output of \(x_s\), \(x_t\) in the \(l\)-th} $
$ \text{\qquad \qquad feature block: \(f_s^l\), \(f_t^l \).} $

\STATE $ \text{\qquad Update model parameters $\theta$ by Eq.~(\ref{eq1}).} $  
\STATE $ \textbf{End For} $
\STATE Return model parameters $\theta$.
\end{algorithmic}
\label{alg1}
\end{algorithm}

\subsubsection{Target-Free CUTI-Domain}

Sometimes, the unauthorized target domain is unknown. In this case,  we cannot directly feed the target domain and CUTI-domain into the network for model training. To solve this, a synthesized target domain could be utilized. For example,  Wang~\etal~\cite{ref9} designed a GAN-based method by freezing parameters to generate synthesized samples in different directions in place of the target domain train set. Although their method is able to generate high-quality synthesized samples, the direction of the generator is specified and there are certain omissions. Huang~\etal~\cite{ref33} designed an adaptive instance normalization (AdaIN) method based on GAN, which can generate synthesized images with a specific style for a content image.

To take full advantage, we add Gaussian noise to AdaIN to obtain synthesized samples with random styles. Finally, we mix the synthesized samples generated by the above two methods (\textit{e.g.}, GAN and AdaIN) to replace the target domain training set and initialize the CUTI-domain. 
The framework is consistent with the target-specified CUTI-domain and the training process is detailed in the Appendix.
During testing, the model is evaluated on source domain test set and other unknown domain with the same task. Our goal is not to design GANs to generate high-quality synthesized images, but to evaluate the ability of CUTI-domain to block illegal feature transfer in the context of synthesized images. We still focus on reducing the feature recognition ability of the model on unauthorized target domains and maintaining it on the authorized source domains.

\begin{table*}[!t]
  \centering
    \caption{The accuracy ($\%$) of target-specified CUTI-domain on digit datasets. The left of ‘\(\Rightarrow\)’ represents the accuracy of the model trained on the source domain dataset with SL, and the right of ‘\(\Rightarrow\)’ is the accuracy of CUTI-domain. CUTI source/target drop represent the average degradation (relative degradation) of the proposed CUTI-domain relative to SL on the source/target domains. NTL source/target drop is calculated from the original paper. The bold numbers indicate the best performance.}
  \resizebox{1\textwidth}{!}{
  \begin{tabular}{ccccccccccccccccc}
    \toprule
    Source/Target &  MT & US & SN & MM & \makecell[c]{CUTI \\ Source \\ Drop$\downarrow$} & \makecell[c]{CUTI \\ Target \\ Drop$\uparrow$} & \makecell[c]{NTL \\ Source \\ Drop$\downarrow$} & \makecell[c]{NTL \\ Target \\ Drop$\uparrow$}\\
    \midrule
    MT & 99.2 $\Rightarrow$  99.1 & 98.0 $\Rightarrow$ \ 6.7 & 38.2 $\Rightarrow$ \ 5.6 & 67.8 $\Rightarrow$ \ 8.7 & \bf 0.10 (0.10\%) & \bf 61.00 (88.56\%) & 1.00 (1.01\%) & 46.57 (75.60\%)\\
    US & 92.6 $\Rightarrow$  10.0 & 99.7 $\Rightarrow$  99.6 & 25.5 $\Rightarrow$ \ 6.8 & 41.2 $\Rightarrow$ \ 8.4 & \bf 0.10 (0.10\%) & \bf 44.70 (80.72\%) & 1.00(1.00\%) & 38.67 (75.55\%)  \\
    SN & 66.7 $\Rightarrow$ \ 9.2 & 70.5 $\Rightarrow$ \ 6.7 & 91.2 $\Rightarrow$  90.9 & 34.6 $\Rightarrow$  10.9 & \bf 0.30 (0.33\%) & \bf 48.33 (81.73\%) & 1.10(1.23\%) & 40.60 (77.25\%) \\
    MM & 98.4 $\Rightarrow$ \ 9.5 & 88.4 $\Rightarrow$ \ 6.8 & 46.3 $\Rightarrow$ \ 7.6 & 95.4 $\Rightarrow$  95.4 & \bf 0.00 (0.00\%) & \bf 69.73 (88.75\%) & 2.10(2.30\%) & 60.10 (76.95\%) \\
    \midrule
    Mean & / & / & / & / & \bf 0.13 (0.13\%) & \bf 55.94 (84.94\%) & 1.30 (1.39\%) & 46.48 (76.34\%) \\
    \bottomrule
  \end{tabular}}

  \label{tab1}
\end{table*}

\begin{figure*}[!t]
  \centering
   \includegraphics[width=0.95\linewidth,trim=30 184 20 184,clip]{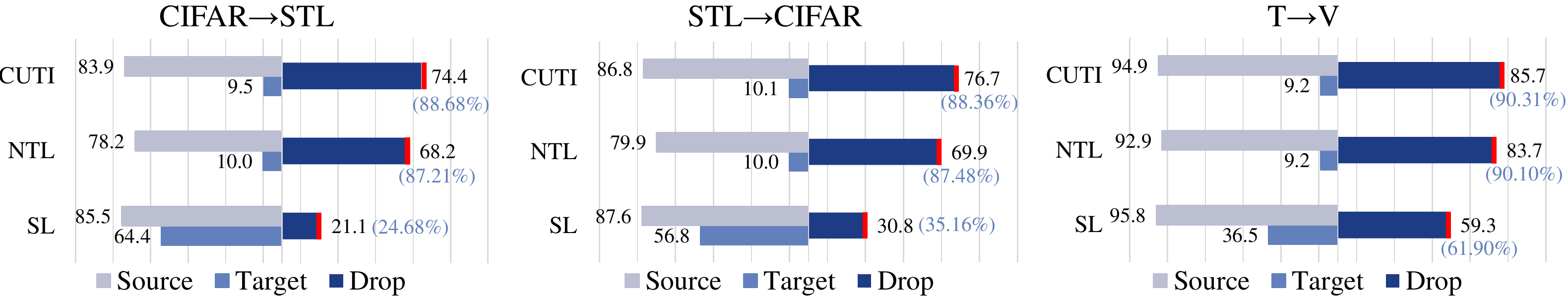}

   \caption{The accuracy ($\%$) of SL, target-specified NTL and target-specified CUTI-domain on CIFAR10, STL10 and VisDA-2017. The left of ‘\(\rightarrow\)’ represents the source domain and the right of ‘\(\rightarrow\)’ is the target domain. The bars with different colors in each subgraph represent the accuracy of the corresponding method in the source domain, the target domain, and the degradation (relative degradation) of the model performance, respectively. The data of NTL is obtained by reproducing its source code.}
   \label{figure4}
\end{figure*}

\begin{table*}[!t]
  \centering
  \caption{The accuracy ($\%$) of ownership verification by SL and CUTI-domain. FTAL~\cite{ref25}, RTAL~\cite{ref25}, EWC~\cite{ref26}, AU~\cite{ref26} and Overwriting are state-of-the-art watermark removal methods. Avg drop represents the average degradation of the test dataset with watermark patch relative to test dataset without a watermark patch. The bold numbers indicate the best performance.}
  \resizebox{1\textwidth}{!}{
  \begin{tabular}{c|cc|ccccc|cc}
    \toprule
    \multirow{3}{*}{\makecell[c]{Source \\ without \\ Patch}} & \multicolumn{2}{c|}{Training Methods} & \multicolumn{5}{c|}{Watermark Removal Approaches on CUTI} & \multicolumn{2}{c}{Avg Drop$\uparrow$}\\
    \cline{2-10}
    & \makecell[c]{SL} & CUTI & FTAL~\cite{ref25} & RTAL~\cite{ref25} & EWC~\cite{ref26} & AU~\cite{ref26} & Overwriting & \multirow{2}{*}{CUTI} & \multirow{2}{*}{NTL}\\
    & \multicolumn{2}{c|}{[Test w/o Watermark ($\%$)]} & \multicolumn{5}{c|}{[Test w/o Watermark ($\%$)]}\\
    \midrule
    MT & 99.0 $/$ 99.3 &  11.3 $/$ 99.1 & \ 9.0 $/$ 100.0 & \ 9.4 $/$  100.0 & \ 9.7 $/$ 100.0 & \ 9.0 $/$  100.0 & \ 9.4 $/$ 96.2 & \bf 89.9 & 88.4\\
    US & 99.8 $/$ 99.8 & \ 7.7 $/$ 99.8 & \ 8.0 $/$ 100.0 & \ 8.7 $/$  100.0 & \ 9.4 $/$ 100.0 & \ 9.4 $/$  100.0 & \ 8.7 $/$ 98.6 & \bf 90.9 & 85.7\\
    SN & 91.3 $/$ 92.3 & \ 9.9 $/$ 92.1 & \ 9.4 $/$  98.3 &  13.9 $/$ \ 97.6 &  10.8 $/$ 100.0 & \ 8.7 $/$  100.0 &  10.4 $/$ 95.8 & \bf 87.7 & 79.0\\
    MM & 96.6 $/$ 96.0 &  16.8 $/$ 96.0 &  14.3 $/$  95.4 &  24.0 $/$ \ 98.6 &  14.6 $/$ 100.0 &  14.6 $/$  100.0 &  14.9 $/$ 95.8 & \bf 81.5 & 77.3\\
    CIFAR & 83.3 $/$ 75.1 &  10.7 $/$ 86.8 &  14.9 $/$  97.9 &  14.9 $/$ \ 93.8 &  14.9 $/$ 100.0 & \ 9.4 $/$ \ 97.2 &  16.7 $/$ 90.3 & \bf 81.7 & 74.6\\
    STL & 87.9 $/$ 93.2 &  22.0 $/$ 88.2 &  20.0 $/$  96.9 &  26.4 $/$ \ 93.8 &  13.9 $/$ 100.0 &  22.9 $/$ \ 94.1 &  21.2 $/$ 89.6 & \bf 74.0 & \bf 74.0\\
    VisDA & 93.6 $/$ 92.2 &  13.1 $/$ 94.1 &  15.0 $/$  95.5 &  20.5 $/$ \ 95.1 &  15.3 $/$ 100.0 &  21.9 $/$ \ 95.1 &  19.4 $/$ 96.2 & \bf 78.0 & 76.8\\
    \midrule
    Mean & / & / & / & / & / & / & / & \bf83.4 & 79.4 \\
    \bottomrule
  \end{tabular}}
  \label{tab2}
\end{table*}

\begin{table*}[!t]
  \centering
  \caption{The accuracy ($\%$) of target-free CUTI-domain on digit datasets. The left of ‘\(\Rightarrow\)’ represents the accuracy of the model trained on the source domain dataset with SL, and the right of ‘\(\Rightarrow\)’ is the accuracy of CUTI-domain. CUTI source/target drop represent the average degradation (relative degradation) of the proposed CUTI-domain relative to SL on the source/target domains. The data of NTL is obtained by reproducing its source code. The bold numbers indicate the best performance.}
  \resizebox{1\textwidth}{!}{
  \begin{tabular}{ccccccccccccccccc}
    \toprule
    Source/Target &  MT & US & SN & MM & \makecell[c]{CUTI \\ Source \\ Drop$\downarrow$} & \makecell[c]{CUTI \\ Target \\ Drop$\uparrow$} & \makecell[c]{NTL \\ Source \\ Drop$\downarrow$} & \makecell[c]{NTL \\ Target \\ Drop$\uparrow$}\\
    \midrule
    MT & 99.2 $\Rightarrow$  98.8 & 98.0 $\Rightarrow$ \ 6.7 & 38.2 $\Rightarrow$ \ 6.7 &  67.8 $\Rightarrow$  13.1 & \bf 0.40 (0.40\%) & \bf 59.17 (85.43\%) & 0.70 (0.71\%) & 57.30 (84.06\%) \\
    US & 92.6 $\Rightarrow$ \ 9.1 & 99.7 $\Rightarrow$  99.7 & 99.1 $\Rightarrow$  25.5 & \ 6.8 $\Rightarrow$ \ 8.5 & \bf 0.60 (0.60\%) & \bf 44.97 (80.96\%) & \bf 0.60 (0.60\%) & 42.90 (74.71\%) \\
    SN & 66.7 $\Rightarrow$  11.9 & 70.5 $\Rightarrow$  14.3 & 91.2 $\Rightarrow$  88.7 &  34.6 $\Rightarrow$  13.6 & \bf 2.50 (2.74\%) & \bf 44.00 (74.19\%) & 3.20 (3.51\%) & 41.53 (64.09\%) \\
    MM & 98.4 $\Rightarrow$  19.6 & 88.4 $\Rightarrow$ \ 6.8 & 46.3 $\Rightarrow$ \ 9.5 &  95.4 $\Rightarrow$  95.1 & \bf 0.30 (0.31\%) & \bf 65.73 (83.96\%) & 2.00 (2.10\%) & 63.03 (77.62\%) \\
    \midrule
    Mean & / & / & / & / & \bf 0.95 (1.02\%) & \bf 53.47 (81.13\%) & 1.63 (1.73\%) & 51.19 (75.12\%) \\
    \bottomrule
  \end{tabular}}
  \label{tab3}
\end{table*}

\section{Experiment}
\subsection{Implementation Details}
We evaluate our method on seven popular DA/DG benchmarks, \textit{i.e.}, MNIST (MT)~\cite{ref34}, USPS (US)~\cite{ref35}, SVHN (SN)~\cite{ref36} and MNIST-M (MM)~\cite{ref37} are commonly used digit datasets, containing ten digits from 0 to 9 extracted from vary scenes; CIFAR10~\cite{refCIFAR10} and STL10~\cite{refSTL10} are all ten-class classification datasets. We follow the procedure of French et al.~\cite{ref39} to process the dataset such that the correspondence between them holds. VisDA-2017~\cite{ref40} is a Synthetic-to-Real dataset containing training (T) and validation (V) sets from 12 categories. Following the general setup, we adopt accuracy ($\%$) as the performance metric of each task. 

For tasks with different complexity, different backbones are adopted to compare with the method proposed by Wang (NTL)~\etal~\cite{ref9}. We used VGG-11~\cite{ref41} for digit datasets, VGG-13~\cite{ref41} for CIFAR10 and STL10, and VGG-19~\cite{ref41} for VisDA. Since VGG contains five feature extractor blocks, \(L\)  is set to 5 in this paper, and CUTI-domain generator is deployed after each max-pooling layer in the block. Pre-trained models are used for all backbones for a fair comparison. The implementation of our comprehensive experiments is based on the public platform Pytorch and an NVIDIA GeForce RTX 3090 GPU with 24GB of memory. The batch size for each domain is set to 32.

\begin{figure*}[!t]
  \centering
   \includegraphics[width=0.95\linewidth,trim=20 184 30 184,clip]{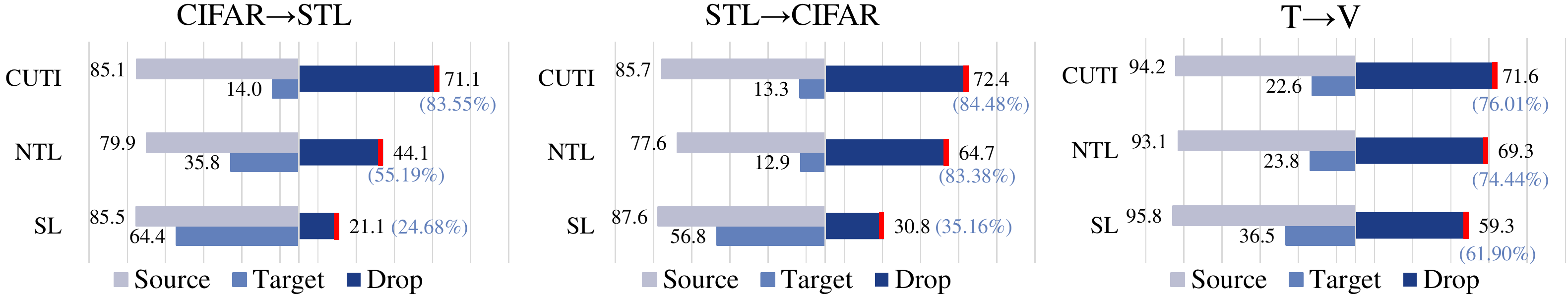}
   \caption{The accuracy ($\%$) of SL, target-free NTL and target-free CUTI-domain on CIFAR10, STL10, and VisDA-2017. The left of ‘\(\rightarrow\)’ represents the source domain and the right of ‘\(\rightarrow\)’ is the target domain. The bars with different colors in each subgraph represent the accuracy of the corresponding method in the source domain, the target domain, and the degradation (relative degradation) of the model performance, respectively. The data of NTL is obtained by reproducing its source code.}
   \label{figure5}
   \vskip -15pt
\end{figure*}

\begin{table*}
  \centering
  \caption{The accuracy ($\%$) of applicability authorization on digit datasets, CIFAR10, STL10 and VisDA-2017. CUTI authorized/other domain represent the average accuracy of CUTI-domain on the authorized/other domain, CUTI drop denote the degradation (relative degradation). The results of NTL are reproduced by its source code. The bold numbers indicate the best performance.}
  \resizebox{1.\textwidth}{!}{
\begin{tabular}{c|cccccccc|cccccc}
\hline
\multirow{3}{*}{\begin{tabular}[c]{@{}c@{}}Source\\ with \\ Path\end{tabular}} & \multicolumn{4}{c|}{\multirow{2}{*}{Test with Path(\%)}} & \multicolumn{4}{c|}{\multirow{2}{*}{Test without Path(\%)}} & \multicolumn{1}{c}{\multirow{3}{*}{\begin{tabular}[c]{@{}c@{}}CUTI\\ Authorized\\ Domain$\uparrow$\end{tabular}}} & \multicolumn{1}{c}{\multirow{3}{*}{\begin{tabular}[c]{@{}c@{}}CUTI\\ Other\\ Domain$\downarrow$\end{tabular}}} & \multicolumn{1}{c}{\multirow{3}{*}{\begin{tabular}[c]{@{}c@{}}CUTI\\ Drop\\ $\uparrow$\end{tabular}}} & \multicolumn{1}{c}{\multirow{3}{*}{\begin{tabular}[c]{@{}c@{}}NTL\\ Authorized\\ Domain$\uparrow$\end{tabular}}} & \multicolumn{1}{c}{\multirow{3}{*}{\begin{tabular}[c]{@{}c@{}}NTL\\ Other\\ Domain$\downarrow$\end{tabular}}} & \multirow{3}{*}{\begin{tabular}[c]{@{}c@{}}NTL\\ Drop\\ $\uparrow$\end{tabular}} \\
 & \multicolumn{4}{c|}{} & \multicolumn{4}{c|}{} & \multicolumn{1}{c}{} & \multicolumn{1}{c}{} & \multicolumn{1}{c}{} & \multicolumn{1}{c}{} & \multicolumn{1}{c}{} &  \\ 
\cline{2-9}
 & MT & US & SN & \multicolumn{1}{c|}{MM} & MT & US & SN & MM & \multicolumn{1}{c}{} & \multicolumn{1}{c}{} & \multicolumn{1}{c}{} & \multicolumn{1}{c}{} & \multicolumn{1}{c}{} &  \\ 
\hline
MT & 100.0 & 14.3 & 17.6 & \multicolumn{1}{c|}{12.9} & 10.3 & 8.6 & 18.3 & 14.1 & \bf 100.0 & \bf 13.7 & \bf 86.27(86.27\%) & 99.8 & 14.5 & 85.31(85.49\%) \\
US & 9.6 & 99.2 & 14.9 & \multicolumn{1}{c|}{10.7} & 10.2 & 6.7 & 8.6 & 10.6 & \bf 99.2 & \bf 10.2 & \bf 89.01(89.73\%) & 98.5 & 13.3 & 85.20(86.50\%) \\
SN & 10.8 & 13.5 & 99.1 & \multicolumn{1}{c|}{23.0} & 9.6 & 9.2 & 17.7 & 8.9 & 99.1 & \bf 13.2 & \bf 85.86(86.64\%) & \bf 99.3 & 15.8 & 83.51(84.10\%) \\
MM & 8.9 & 9.1 & 15.0 & \multicolumn{1}{c|}{99.5} & 10.0 & 9.1 & 11.9 & 25.9 & \bf 99.5 & \bf 12.8 & \bf 86.66(87.09\%) & \bf 99.5 & 14.0 & 85.49(85.92\%) \\ 
\hline
 & \multicolumn{2}{c}{CIFAR} & \multicolumn{2}{c|}{STL} & \multicolumn{2}{c}{CIFAR} & \multicolumn{2}{c|}{STL} & \multicolumn{6}{c}{/} \\ 
\hline
CIFAR & \multicolumn{2}{c}{97.9} & \multicolumn{2}{c|}{42.5} & \multicolumn{2}{c}{13.1} & \multicolumn{2}{c|}{11.6} & \bf 97.9 & \bf 22.4 & \bf 75.50(77.12\%) & 97.5 & 24.6 & 72.90(74.77\%) \\
STL & \multicolumn{2}{c}{29.0} & \multicolumn{2}{c|}{99.9} & \multicolumn{2}{c}{13.3} & \multicolumn{2}{c|}{15.2} & \bf 99.9 & \bf 19.2 & \bf 80.73(80.81\%) & 98.6 & 20.5 & 78.10(79.21\%) \\ 
\hline
 & \multicolumn{2}{c}{T} & \multicolumn{2}{c|}{V} & \multicolumn{2}{c}{T} & \multicolumn{2}{c|}{V} & \multicolumn{6}{c}{/} \\ 
\hline
T & \multicolumn{2}{c}{100.0} & \multicolumn{2}{c|}{22.9} & \multicolumn{2}{c}{20.3} & \multicolumn{2}{c|}{10.1} & \bf 100.0 & \bf 17.8 & \bf 82.23(82.23\%) & \bf 100.0 & 23.3 & 76.70(76.70\%) \\ 
\hline
Mean & \multicolumn{8}{c|}{/} & \bf 99.4 & \bf 15.5 & \bf 83.75(84.27\%) & 99.0 & 18.0 & 81.03(81.81\%) \\ 
\hline
\end{tabular}
}
  \vskip -10pt
  \label{tab4}
\end{table*}

\subsection{Result of Target-Specified CUTI-Domain}
We selected one as the source domain and one as the target domain from the digital datasets of four different domains, and constructed 16 transfer tasks, as shown in Table~\ref{tab1}. The left of \(\Rightarrow\) represents the accuracy of the model trained on the source domain dataset using supervised learning (SL), and the right of \(\Rightarrow\) is the accuracy of CUTI-domain. CUTI source/target drops represent the drop (relative drop) of the proposed CUTI-domain in the source and target domains, respectively. As can be seen, the average drop of CUTI-domain on the target and source domains is 55.94 (84.94\%) and 0.13 (0.13\%), respectively. The last two columns represent the average performance degradation of NTL with values of 46.48 (76.34\%) and 1.30 (1.39\%), respectively. Compared with NTL, the decline of CUTI-domain in the target domain is higher, and the negative impact on the source domain is smaller. It can be inferred that CUTI-domain can better reduce the sample recognition ability of the model for the target domain, and the decreases on the source domain are slight. 

Fig.~\ref{figure4} shows the results on CIFAR10 \(\rightarrow\) STL10, STL10 \(\rightarrow\) CIFAR10 and T \(\rightarrow\) V. The bars with different colors in each subgraph represent the accuracy of the corresponding method in the source domain, the accuracy in the target domain, and the degradation (relative degradation) of the model performance, respectively. Where the results of NTL are reproduced by its source code to obtain comparable experimental data. SL has the largest generalization region, resulting in the highest classification accuracy on the target domain. By blocking the pathway with model locker, we observe successful target domain reduction in accuracy for NTL and CUTI-domain, with higher degradation than SL. Meanwhile, regardless of the task, the degradation of CUTI-domain is higher than that of NTL, indicating that CUTI-domain can better compress the generalization region of the model.

\subsection{Result of Ownership Verification}
In this section, we conduct ownership verification of model by triggering classification errors. Specifically, we add a regular backdoor-based model watermark patch on the authorized source domain dataset followed by NTL~\cite{ref9}, and treat the processed source domain as the new unauthorized target domain. The model classification accuracy of SL and CUTI-domain on the authorized source domain without watermark patch and the unauthorized target domain with watermark patch are shown in Table~\ref{tab2}. 
As shown in the second and third columns in the table, for SL, there is little difference in the accuracy before and after embedding the watermark patch, so the model is not sensitive to the watermark patch. While for CUTI-domain, after watermark patch embedding, the accuracy on unauthorized target domain is greatly reduced, and this difference in performance can be used to verify the ownership of the model. 
In addition, we also test the robustness of CUTI-domain using FTAL~\cite{ref25}, RTAL~\cite{ref25}, EWC~\cite{ref26}, AU~\cite{ref26} and watermark overwriting, which are state-of-the-art model watermark removal methods. For a fair comparison, the settings of these watermark removal methods are all consistent with NTL. The last two columns of Table~\ref{tab2} are the drop in accuracy for watermarked versus un-watermarked data. It can be observed that both CUTI-domain and NTL can effectively resist the attack of watermark removal method, and the performance of CUTI-domain is about 4\% higher.

\subsection{Result of Target-free CUTI-Domain}
As detailed in Section 3.3.2, when the target domain is unknown, we use synthesized samples to replace the target domain training set, and use other unknown domain datasets as the target domain test set, as shown in Table~\ref{tab3} and Fig.~\ref{figure5}. For a fair comparison, the data of NTL is obtained by reproducing its source code. It can be observed that the average drop of CUTI-domain on the target domain is higher than that of NTL, and the drop on the source domain is lower, indicating better IP protection ability of our proposed CUTI-domain on unauthorized domains.

Fig.~\ref{figure5} shows the results for CIFAR10 \(\rightarrow\) STL10, STL10 \(\rightarrow\) CIFAR10 and T \(\rightarrow\) V.
Where the results of NTL are reproduced by its source code to obtain comparable experimental data. Consistent with the previous text, the degradation of CUTI-domain and NTL are higher than SL, and CUTI-domain is the highest, which implies that CUTI-domain can better compress the generalization region of the model. Meanwhile, considering the complexity of the VisDA-2017 dataset, it is more difficult to extract representative features to build CUTI-domain, so the drop of T \(\rightarrow\) V of CUTI-domain is only slightly higher than NTL.

\subsection{Result of Applicability Authorization}
In this section, we validate the applicability of model by restricting its generalization ability to the authorized source domain. Specifically, similar to section 4.3, we add an authorized watermark patch on the source domain as a new authorized source domain training set. Then the original source domain, synthesized samples, and synthesized samples with authorized watermark patches are mixed as the unauthorized target domain training set. During testing, other unknown domains are used as the test set. The experimental results of CUTI-domain are shown in Table~\ref{tab4}, where the results of NTL are reproduced by its source code. 
It can be observed that the model performs better on the source domain with authorized watermark patch, but performs poorly on other unknown domains with or without watermark patch. This is consistent with our expectation that the generalization ability of the model is restricted to the source domain with the authorized watermark patch. 
Meanwhile, the average drop rate of our proposed CUTI-domain is 83.75 (84.27\%), which is higher than NTL with 81.03 (81.81\%). This is mainly because NTL directly utilizes the limited features of the source and target domains for distance maximization, while we construct a CUTI-domain with infinite samples similar to the source domain, whose generalization boundary is more compact, and thus the model IP protection ability is stronger.

\subsection{Ablation Study}
\begin{figure}[!t]
  \centering
   \includegraphics[width=1\linewidth,trim=190 141 190 124,clip]{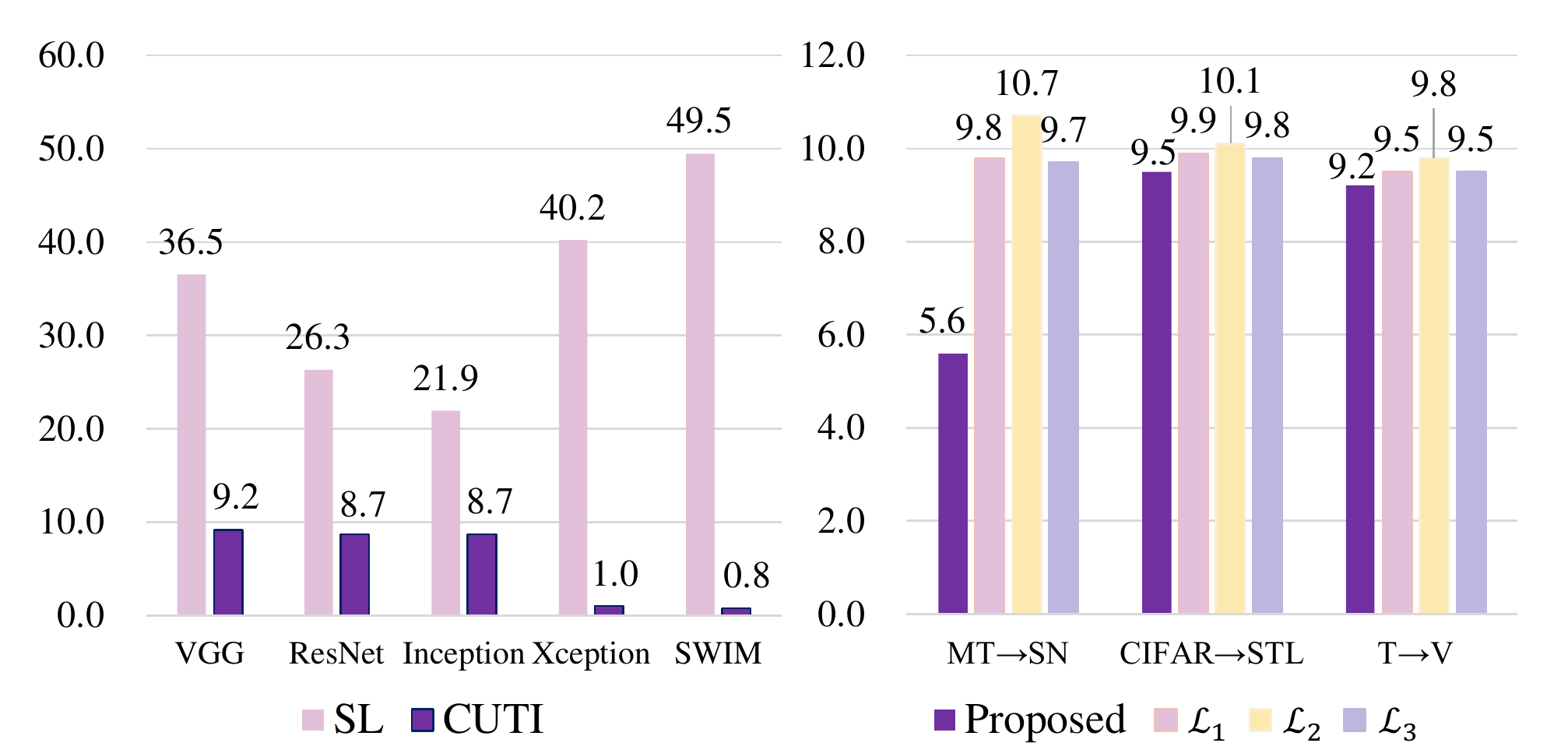}
   \caption{\textbf{Left:} The accuracy ($\%$) of SL and target-specified CUTI-domain combined with different backbones on VisDA-2017. \textbf{Right:} The accuracy ($\%$) of SL and target-specified CUTI-domain with different loss functions on three random tasks.}
   \label{figure6}
   \vskip -15pt
\end{figure}

\textbf{Backbone:}
In this section, we verify the ability of IP protection of CUTI-domain combined with other backbones on the VisDA-2017 dataset. As shown in the left of Fig.~\ref{figure6}, compared with SL, CUTI-domain can further reduce the recognition ability on the target domain when implemented with VGG-19~\cite{ref41}, ResNet-34~\cite{ref51}, Inceptionv3~\cite{ref52}, Xception~\cite{ref53} and SWIM~\cite{ref54}, Meanwhile, the accuracy of the CUTI-domain on target domain is lower when combined with Xception~\cite{ref53} and SWIM~\cite{ref54}, since they have stronger feature extraction ability on complex datasets than VGG-19~\cite{ref41}, ResNet-34~\cite{ref51} and Inceptionv3~\cite{ref52}, lead to building a CUTI-domain that is more similar to the source domain, thus better compacting the model performance within the source domain, implying stronger model IP protection ability.

\textbf{Loss Function:}
We further explore the contribution of various parts of our proposed alternative loss function \(\cal L\) in Eq.~(\ref{eq1}). The variants of \(\cal L\) are designed as:
\begin{align}
{\cal L}_1 & = KL(p_s||y_s) - KL(p_t||y_t), \\
{\cal L}_2 & = KL(p_s||y_s) - KL(p_i||y_i), \\
{\cal L}_3 & = KL(p_s||y_s) - KL(p_t||y_t) - KL(p_i||y_i).
\end{align}
We validate the performance of different loss function variants on three random tasks, as shown in the right of Fig.~\ref{figure6}. 
Due to the different complexity of the datasets, the difficulties of feature extraction and CUTI-domain construction are varying. On datasets with simple features (\textit{i.e.}, MT \(\rightarrow\) SN), \(\cal L\) performs better, while on slightly more complex datasets, \(\cal L\) only slightly outperforms other loss functions. It can be seen that the accuracy scores of different loss function variants on the target domain are relatively close, and \(\cal L\) has the lowest accuracy on each dataset, implying the validity of our proposed alternative loss function \(\cal L\).

\section{Conclusion}
In the field of artificial intelligence, protecting well-trained models as a form of IP poses challenges. To address this issue, we propose a novel CUTI-domain that acts as a barrier to constrain model performance to authorized domains. Our approach involves creating an isolation domain with features similar to those in the authorized domain, effectively blocking the model's pathway between authorized and unauthorized domains and leading to recognition failure on unauthorized domains with new private style features. We also offer two versions of the CUTI-domain, \textit{e.g.}, target-specified and target-free, depending on whether the unauthorized domain is known. Our experimental results on seven popular cross-domain datasets demonstrate the efficacy of our lightweight, plug-and-play CUTI-domain module. We hope that our work could promote the research of model IP protection and security, which should be taken seriously in real-world applications. 

\small{\noindent \textbf{Acknowledge:} This work was supported by the National Natural Science Foundation of China (Nos. 62136004, 62276130), the Key Research and Development Plan of Jiangsu Province (No. BE2022842), and Huazhu Fu's A*STAR Central Research Fund and Career Development Fund (C222812010).}
%-------------------------------------------------------------------------

%-------------------------------------------------------------------------

%%%%%%%%% REFERENCES
{\small
\bibliographystyle{ieee_fullname}
\bibliography{CUTI_finial}
}
\end{document}